\documentclass[letterpaper, 10 pt, conference]{ieeeconf}  

\IEEEoverridecommandlockouts                              

\usepackage{graphics} 
\usepackage{epsfig} 
\usepackage{mathptmx} 
\usepackage{times} 
\usepackage{amsmath} 
\usepackage{amssymb}  
\usepackage{hyperref}
\usepackage{tikz}
\usepackage{comment}
\usepackage{color}
\usepackage{multirow}

\DeclareMathOperator*{\argmin}{arg\,min}

\title{Prior Based Online Lane Graph Extraction from Single Onboard Camera Image}

\author{Yigit Baran Can$^{1}$ \quad Alexander Liniger$^{1}$\quad  Danda Paudel$^{1}$\quad  Luc~Van~Gool$^{1,2}$ \thanks{
$^{1}$Computer Vision Laboratory, ETH Zurich, Switzerland, {\tt\small\{yigit.can,alex.liniger,paudel, vangool\}@vision.ee.ethz.ch} $^{2}$KU Leuven, Belgium. 
}}

\begin{document}

\maketitle
\thispagestyle{empty}
\pagestyle{empty}


\begin{abstract}

The local road network information is essential for autonomous navigation. This information is commonly obtained from offline HD-Maps in terms of lane graphs. However, the local road network at a given moment can be drastically different than the one given in the offline maps; due to construction works, accidents etc. Moreover, the autonomous vehicle might be at a location not covered in the offline HD-Map. Thus, online estimation of the lane graph is crucial for widespread and reliable autonomous navigation. In this work, we tackle online Bird's-Eye-View lane graph extraction from a single onboard camera image. We propose to use prior information to increase quality of the estimations. The prior is extracted from the dataset through a transformer based Wasserstein Autoencoder. The autoencoder is then used to enhance the initial lane graph estimates. This is done through optimization of the latent space vector. The optimization encourages the lane graph estimation to be logical by discouraging it to diverge from the prior distribution. We test the method on two benchmark  datasets, NuScenes and Argoverse. The results show that the proposed method significantly improves the performance compared to state-of-the-art methods. Code: \url{https://github.com/ybarancan/lanewae}.
\end{abstract}

\section{Introduction}

Autonomous navigation tasks require accurate information about the surrounding environment. In the field of autonomous driving, this translates to the estimation of the local road network. The local road network forms the foundation for several crucial downstream tasks such as predicting the motion of agents \cite{cui2019multimodal, hong2019rules, rella2021decoder, zaech2020action} and to plan the ego-motion \cite{DBLP:conf/rss/BansalKO19, chen2020learning}. Commonly the local road network is obtained from an offline generated HD-map, possibly in conjunction with an onboard perception module~\cite{jaritz20202d,ma2019exploiting,casas2021mp3}. However, depending on offline HD-maps can result in failures, when the networked changes over time. Events such as accidents, construction work, temporary re-routings require an online method to predict the local road network. Moreover, dependence on offline maps severely limits the scalability of autonomous driving by restricting the application to the areas with offline map coverage.

\begin{figure}
    \centering
    \includegraphics[width=\linewidth]{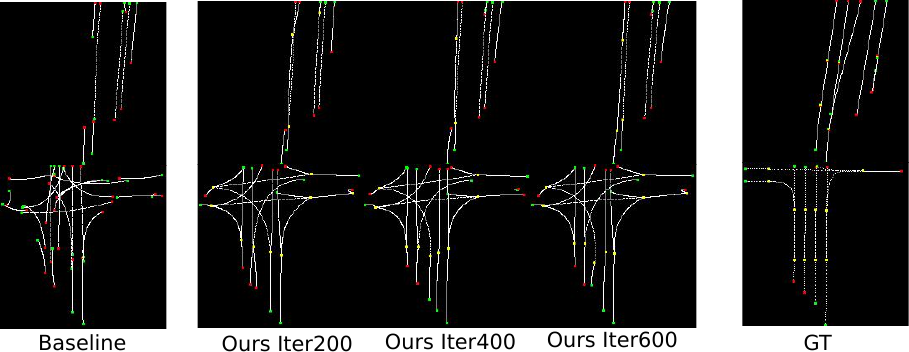}
    \vspace{-1.5em}
    \caption{
    Left to right: the lane graphs estimated using the state-of-the-art method \cite{Can_2021_ICCV}, the proposed method, and the ground truth (GT). Our method takes an initial estimate (baseline method's estimate) and performs latent space optimization to improve the initial results. 
    }
    \label{fig:rebut}
    \vspace{-1em}
\end{figure}

Online mapping for autonomous navigation requires the understanding of the surrounding in Bird's-Eye-View (BEV) which poses additional problems to the traditional scene understanding tasks where the target representation lives in the same plane as the input, which is usually the image plane. It has been shown that producing the output representation directly in the target plane is superior to 2-step approach of image level understanding followed by projecting the results to the gound~\cite{DBLP:conf/cvpr/RoddickC20,can2020understanding,wu2020motionnet,philion2020lift, zhou2022cross}. Most of the methods in literature, however, do not provide the local road network, but a segmentation on the BEV. Some methods extract lane boundaries instead of BEV segmentation~\cite{DBLP:conf/ivs/NevenBGPG18, DBLP:conf/iccvw/GansbekeBNPG19} while lane graphs from highways were extracted using LiDAR reflection data~\cite{homayounfar2019dagmapper}. Recently~\cite{Can_2021_ICCV} proposed a method that can predict road networks in urban scenarios with intersections from a single image. In this paper, we are interested in the same setting. However, we show that the our method can significantly improve the performance by performing online optimization over the manifold of the output space. The proposed method produces a structure that is identical to~\cite{Can_2021_ICCV}.

Similar to~\cite{Can_2021_ICCV}, we represent the local road network as a directed graph. The vertices of the graph represent the centerline segments while the edges indicate the connectivity of these segments. The centerline segments are formulated as Bezier curves. We are interested in improving the performance of this representation, especially connectivity, through generative models. We propose a novel transformer based Wasserstein autoencoder~\cite{tolstikhin2018wasserstein} architecture that can process a lane graph. The latent space of this autoencoder is used to improve the initial estimate that is fed to the encoder. Specifically, at test time, a lane graph estimate is obtained given an input image using the discriminative method of~\cite{Can_2021_ICCV}. This estimate is fed to the trained encoder to obtain a latent code. Then, optimization is carried out on the latent code to minimize a novel loss function. The latent vector obtained through this optimization is then decoded using the autoencoder decoder to obtain the final lane graph. The final lane graph has superior performance across all metrics, especially connectivity. Moreover, the resulting lane graph is much more realistic then the initial estimates fed to the autoencoder. Another benefit of the proposed framework is the ability to estimate uncertainty for test samples. Autoencoder architectures have been widely used in literature in different fields for anomaly detection \cite{medel2016anomaly, malhotra2016lstm, eiteneuer2019dimensionality}. In this paper, we use the reconstruction error at the end of iterative latent code optimization as quantification of uncertainty. The experiments show that the uncertainty estimates are highly correlated with the performance metrics, validating the proposed method's ability for uncertainty estimation.

Our major contributions can be summarized as follows:
\begin{enumerate}
\setlength{\itemsep}{0pt}
\setlength{\parskip}{0pt}
\item We propose a novel generative architecture that processes lane graphs. 
 
\item The online refinement  of the lane graphs using a novel loss and latent space optimization  is also proposed. The proposed refinement uses our lane graph generative method. 
\item Proposed framework provides a straightforward and computationally efficient uncertainty estimation method through utilization of reconstruction loss arising from the optimized latent code
\item The results obtained by our method are superior to the compared baseline and the state-of-the-art methods, especially in the aspect of connectivity.

\end{enumerate}

\section{Related Works}

Most methods in literature that output lane graphs use aerial images or aggregated sensor data to produce offline HD-Maps. Online methods, on the other hand, can only produce lane boundaries or BEV semantic segmentation. These methods use onboard camera images as well as possibly onboard LiDAR data. 

The earlier works on road network extraction focus on aerial images~\cite{auclair1999survey, richards1999remote} as the input data. This approach has been extensively studied in the literature and more sophisticated and effective methods have been proposed~\cite{batra2019improved,sun2019leveraging,ventura2018iterative}. While aerial image based road network extraction is useful, the resulting lane graph is coarse and not enough for planning tasks. Going beyond road networks, HD-Maps provide a much richer representation of the traffic scene. Most methods in literature construct HD-Maps offline from 2D and 3D information, sometimes aggregated over time~\cite{liang2019convolutional, homayounfar2018hierarchical,liang2018end}. These methods require dense 3D point clouds. Moreover, the use of the offline generated HD-Maps still requires accurate localization of the autonomous agent since these maps are in a canonical frame.

A very relevant line of research for us is lane estimation, especially from onboard monocular camera~\cite{DBLP:conf/ivs/NevenBGPG18, DBLP:conf/iccvw/GansbekeBNPG19,DBLP:conf/cvpr/HomayounfarMLU18}. These methods are restricted to simple cases such as highways and suffer when there are intersections or roundabouts. Because of its advantages and practicality for downstream tasks, representing the scene in BEV has been the preferred approach in the literature. While some methods focus on using only camera images~\cite{DBLP:conf/cvpr/RoddickC20,philion2020lift,can2020understanding}, some other methods combine LiDAR information~\cite{pan2020cross,hendy2020fishing}. None of these methods produce a structured output that can easily be used for downstream tasks. The methods and baselines proposed in~\cite{Can_2021_ICCV} are the first ones to produce a lane graph from a single image. These methods could estimate the centerlines and their connectivity. The centerlines encode the traffic direction as well as producing a complete representation for the road network. However, their work fails to produce realistic and accurate lane graphs in challenging scenarios. The connectivity estimates are especially noisy which results in incorrect merging of the centerlines. 

The proposed framework focuses on online scene understanding and identifying the mistakes made by the mistakes is crucial for safety critical downstream tasks. To this end, we use uncertainty estimation through the reconstruction error of the autoencoder. A popular approach to uncertainty estimation is through the Bayesian approximation by Monte-Carlo dropout \cite{srivastava2014dropout, gal2016dropout} and it has been used in different tasks and fields \cite{kendall2018multi, can2018learning, kendall2015bayesian}. However, this approach requires multiple forward passes at test time to form a Monte-Carlo approximation to posterior distribution. Another approach to uncertainty estimation utilizes the autoencoders. The test-time reconstruction loss of the autoencoder has been used as an estimate for the model uncertainty \cite{medel2016anomaly, malhotra2016lstm, eiteneuer2019dimensionality}. Different than dropout based methods, this approach does not output a distribution but it also does not require additional forward passes.     

Optimizing the latent space of the variational autoencoders have been explored in the literature. Given a set of rules or target functions, \cite{griffiths2020constrained, liu2020chance} formulate a Bayesian optimization problem to enforce specific constraints on the VAE generated samples. In a similar fashion, \cite{gomez2018automatic, weininger1988smiles} show that valid chemical molecules can be generated from a VAE by optimizing the latent space. Another line of research aims to incorporate the uncertainty into the latent optimization problem by proposing the use of an auxiliary network that predicts the value of the objective function given the latent representation \cite{notin2021improving}. All these methods, however, focus on the task of generation whereas we propose a method that can handle a directed acyclic graph (DAG) and produce refined and realistic lane graphs.

\section{Method}

\begin{figure*}
    \centering
    \includegraphics[width=.9\linewidth]{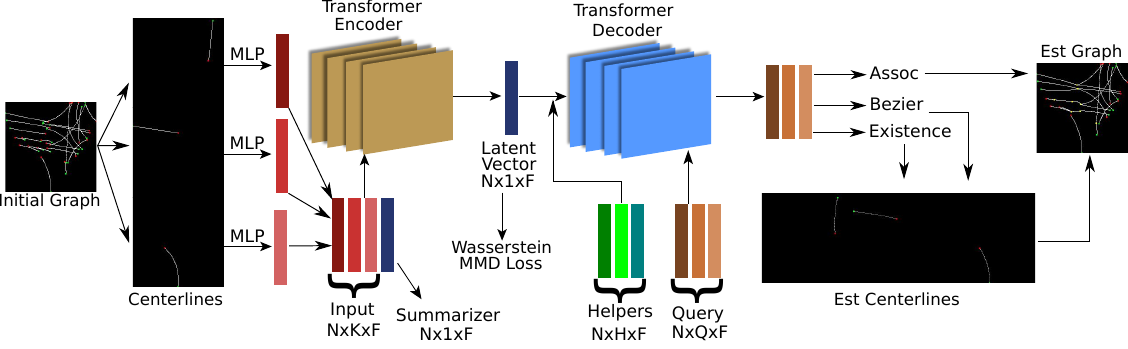}
    \caption{The individual centerlines are converted to feature vectors which are processed together with the summarizer vector to produce the latent vector. The latent vector is then combined with the helper vectors to form the input sequence to the decoder. The learnt query vectors attend the input sequence in the decoder and produce the lane graph estimates.
    }
    \label{fig:transformer}
    \vspace{-1em}
\end{figure*}

\subsection{Lane Graph Representation}
\label{lanegraph}

Lane graphs are directed graphs that represent the road network taking the traffic direction into consideration. The vertices of the graph represent the centerlines and the edges encode the ``connectivity'' of the centerlines. In mathematical terms, the graph $G(V, E)$ has $V$ as the vertices and the edges $E \subseteq \{(x,y)\; |\; (x,y) \in V^2\}$. The incidence matrix $A$ can represent the edges. A centerline $x$ is connected to another centerline $y$, i.e. $(x,y) \in E$ if and only if the centerline $y$'s starting point is the same as the end point of the centerline $x$. This means $A[x,y] = 1$ if the centerlines $x$ and $y$ are connected. Following~\cite{Can_2021_ICCV}, we represent the centerlines, vertices of the graph $G$, with Bezier curves.

\subsection{Motivation}

Generative models have been used to improve the performance of discriminative methods. Some approaches use generative models to produce synthetic datasets on which discriminative models are trained, in order to solve the domain gap problem~\cite{DBLP:journals/corr/abs-1712-00479, DBLP:conf/cvpr/VuJBCP19, DBLP:conf/eccv/ZouYKW18, DBLP:conf/iccv/ChoiKK19}. A similar approach has also been adopted recently to address the problem of semi-supervised learning. A GAN was used to generate new samples from a very limited dataset to enhance the performance of the downstream network~\cite{DBLP:conf/cvpr/ZhangLGYLB0F21}. Another approach that is used to boost performance of discriminative models is to utilize the representations learnt by the generative models in the discriminative model~\cite{galeev2021learning, DBLP:conf/nips/JaakkolaH98, DBLP:conf/nips/RainaSNM03}. These examples clearly indicate that the latent space of the generative models hold useful information which can be utilized by other complimentary networks. In this work, we directly use the latent space of a specially designed transformer based Wasserstein Autoencoder (WAE)~\cite{tolstikhin2018wasserstein} to extract local road network information.

\subsection{Formulation}

Our formulation has an image independent prior part and an image dependent estimation part. Let us represent an input image with $X$ and the correct corresponding lane graph with $\overline{Y}$. Then, our estimate lane graph given the input image $X$ is denoted by $\hat{Y}$. Moreover, let us define a function $Y'=T(X)$ that takes an input image and outputs a lane graph $Y'$. Then, $P(Y) = P_P(Y)*P_O(Y, T(X))$ gives the probability function for a lane graph $Y$ given an image $X$ where $P_P$ is the prior distribution and the $P_O$ is the observation based posterior distribution. Now, let us introduce a variable $Z$ where there exists functions $Z=F(Y')$ and $Y=G(Z)$. In other words, function $F$ maps a given lane graph to a low dimensional vector $Z$ while function $G$ maps this vector into another lane graph $Y$. 

While it is difficult to apply a probability distribution directly on the lane graph $Y$, through functions $F$ and $G$ we can instead define the prior distribution over $Z$. Then, $P_P(Y) = \int_{Z}\delta[G(Z)=Y]P(Z)dZ$ where $\delta[G(Z)=Y]$ is Kronecker delta with $1$ if $G(Z)=Y$ and $0$ otherwise. This integral is very difficult to solve. Thus we will make mild assumptions to significantly simplify it.

Let the training set that is used to train $T(X), F(Y)$ and $G(Z)$ be identical and denoted by $S_T$. Then, we will assume that $T(X)\sim \overline{Y}, \forall X \in S_T$. Note that, this does not imply that $T(X)$ matches the training set distribution but rather only states that the estimates of the function on sample points are close to the true labels. Then, during training, $Z=F(\overline{Y})$ and $Y=G(Z)$ are trained to be as close to $\overline{Y}$ as possible, i.e. $\overline{Y}\sim G(Z), \forall \overline{Y} \in S_T$. This relationship indicates an autoencoder where $F(Y)$ and $G(Z)$ form a bijection between the input lane graph $Y$ and the latent vector $Z$. This lets us approximate the integral $P_P(Y) = \int_{Z}\delta[G(Z)=Y]P(Z)dZ$ with simply a point estimate $P(Z)$ since $\delta[G(Z)=Y]=1$ only holds for one $Z$ value due to bijection. Moreover, we can search for optimal latent vector $Z$ instead of trying to optimize over very complicated $Y$. Then, the overall probability function is given by:
\begin{equation}
   P(Y) = P_O(Y, T(X))*P(Z)\,.
    \label{eq:first-eq}
\end{equation}
We can further modify the expression by using a univariate Normal ($N(0,I)$) for the prior.  As a proxy to maximizing the posterior distribution we minimize a loss function that measures the similarity between the lane graphs $G(Z)$ and $T(X)$. The resulting optimization problem can, then, simply be written as a minimization,
\begin{equation}
    \argmin_Z \quad L_M(G(Z), T(X)) + \alpha||Z||_2  \,,
    \label{eq:second-eq}
\end{equation}
where $\alpha$ is a trade-off parameter and $L_M$ measures the similarity between lane graphs. Given the desired latent vector $Z$, we can obtain the lane graph estimate by $\hat{Y} = G(\hat{Z})$.

The above argument shows that the autoencoder will learn the training set distribution by using the estimates from a trained point estimator. Given the test distribution follows the training distribution, i.e. there is no domain shift, then the optimization given in \eqref{eq:second-eq} can be used to sample from the test set. There is one missing point in our formulation and that is the fact that the latent variable $Z$ follows a univariate Normal distribution. We achieve this by applying Wasserstein regularization on $Z$ similar to~\cite{tolstikhin2018wasserstein}. The details are in Section~\ref{autoencoder}.    

\begin{figure}
    \centering
    \includegraphics[width=\linewidth]{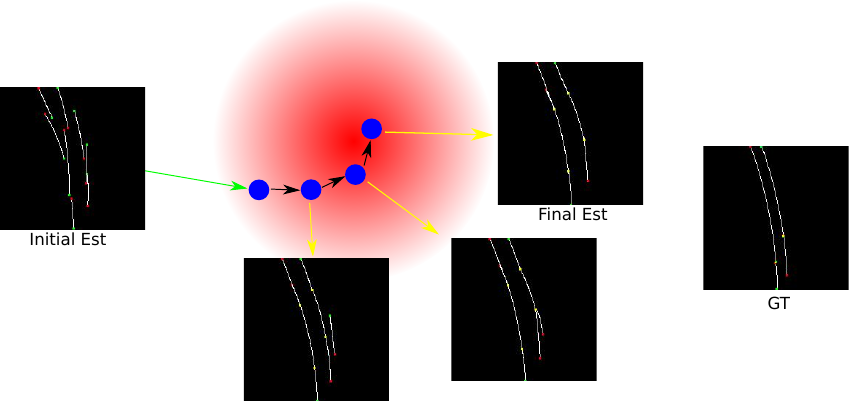}
    \caption{Optimization in latent space minimizes a loss function over the latent code where that latent code is pulled towards the higher prior probability while reproducing the initial estimate. The resulting lane graph is more realistic and accurate than the initial estimate.
    }
    \label{fig:opt}
\end{figure}

\subsection{Autoencoder}
\label{autoencoder}

The input and output of the autoencoder are lane graphs. As explained in Section \ref{lanegraph}, we represent the lane graphs as a directed graph with the vertices being Bezier control points. In order to feed a lane graph into a neural network, we only consider the vertices of the lane graph. Since there is no particular order among the vertices, we can use a transformer based architecture. Let us represent the number of Bezier control points with $B$. Then, we convert $B$ 2-dimensional control points into a flat vector of $2B$ dimensions. This is used by an MLP to produce a feature vector. These feature vectors form our input sequence. 

We need to have a latent space parameterized by a latent vector but the standard transformer does not allow for such a latent vector. Therefore, we have a learnt \emph{summarizer} vector. The summarizer vector is processed together with the input sequence. The decoder of the transformer has only access to the summarizer vector and not the input sequence. In this way, we create a bottleneck that can be used as a latent space. Specifically, the value of the summarizer vector at the end of the encoder, or the beginning of the decoder, is the latent vector $Z$. Throughout the encoder, the summarizer vector attends to the input sequence. This single vector is used by the decoder reproduce the input sequence. 

Given a method for producing a latent vector from an unordered input sequence, we can turn our attention to decode the latent vector into the estimated lane graph. While we do not use the edges of the latent graph, the connectivity information, as input, we estimate it in our output. The idea is that the input centerline segments should be enough for the autoencoder to produce the connectivity since the connectivity is defined by the endpoints of the centerline curves. In order to estimate the lane graph, our method uses learnt query vectors. Each query vector represents a centerline segment (vertex of the lane graph). Each query vector outputs the probability of existence, the values for its Bezier control points and an association feature vector that is used to estimate connectivity, similar to~\cite{Can_2021_ICCV}. 

In order for the formulation given in \eqref{eq:second-eq} to be useful, the latent space of the autoencoder formed by the encoder $F$ and the decoder $G$, should have a univariate Normal distribution. We opt to use a Wasserstein formulation similar to~\cite{tolstikhin2018wasserstein}. WAE penalizes the Wasserstein distance between the sampled latent vectors and the prior distribution, in our case the univariate Normal, to regularize the latent space. Specifically, we use the maximum mean discrepancy (MMD) approach. In this approach, we sample from the desired prior distribution (univariate Normal) and obtain the MMD value between the Normal samples and the latent vector produced by the encoder. The MMD distance is then simply treated as another component of the total loss function which is minimized through gradient descent. Thus, the total loss function to train the autoencoder is ${L_{t} = L_{CE}^{existence} + \beta L_1 + \gamma L_{CE}^{connect} + \theta L_{MMD}}$, where $L_{CE}$ is the cross-entropy loss and $L_{MMD}$ is Wasserstein loss. Existence cross entropy and $L_1$ are applied on matched centerlines while $L_{CE}^{connect}$ is calculated with binary permutations of the estimated centerlines. 

In our experiments, we observed that the transformer decoder cannot produce accurate representations from only a single latent vector. Therefore, we use additional learnt vectors that are combined with the latent vector to form the input sequence. Thus, if there are $H$ helper vectors, the sequence that the decoder takes as input becomes of length $H+1$. The learnt query vectors translate the input sequence into estimated lane graph, see Fig.~\ref{fig:transformer}. We hypothesize that the helper vectors provide a way for the query vectors to avoid the latent vector when needed. This arises from the fact that during attention, the weights are computed through softmax operations. When there is only one source token, i.e. single latent vector, softmax operation is problematic. Through helper vectors, the query vectors can adjust the weight assigned on the latent vector at each recursive iteration step. We observed that the number of helper vectors is not critical as long as it is greater than one.

\begin{figure*}
    \centering
    \includegraphics[width=\linewidth]{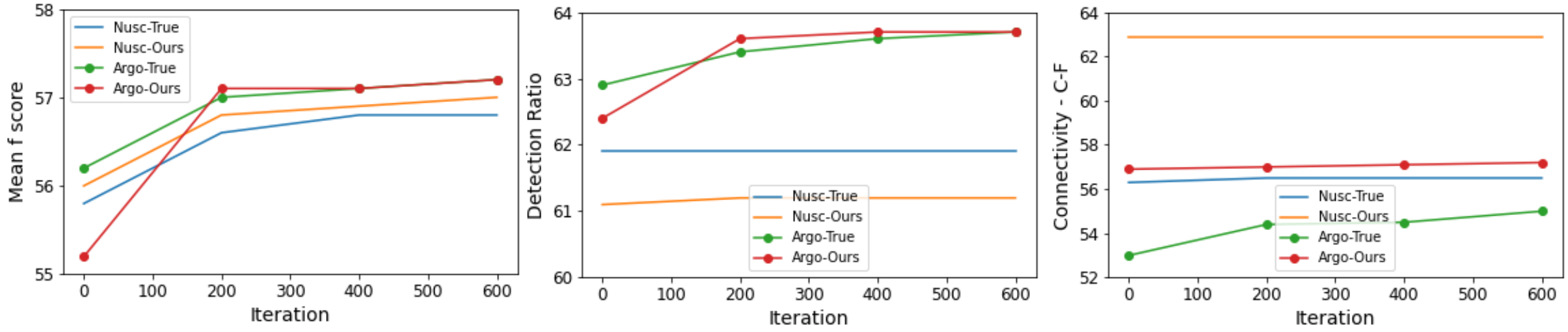}
    \caption{The convergence graphs for the online optimization of the proposed method. The latent space optimization provides substantial benefits. Methods with \textbf{True} suffixes use autoencoders trained with GT lane graphs while \textbf{Ours} use autoencoders trained with the estimated lane graphs. 
    }
    \label{fig:graphs}
\end{figure*}

\subsection{Optimization in Latent Space}

The latent space formed by the WAE is shown to be interpolatable and disentangled~\cite{DBLP:conf/nips/ChakrabartyD21, tolstikhin2018wasserstein,DBLP:journals/corr/abs-1802-03761}, indicating a smooth latent space. This latent space can be used for iterative gradient based optimization techniques. Latent space gradients have been used as a regularizer for the training of generative models~\cite{DBLP:conf/cvpr/KarrasLAHLA20}. In this work, we use the gradients in the latent space to search for a latent code that minimizes a given loss function. 

The loss function $L_M$ given in \eqref{eq:second-eq} is composed of two terms where one measures the average distance between Bezier control points of the centerlines and the other measures the probability of the centerlines existing. Specifically, $L_{m} = L_{CE} + \lambda L_1$, where $L_{CE}$ is the cross-entropy loss on the detection/class probability, and an $L_1$ loss is applied on the Bezier control points. Note that, these losses are applied on the corresponding centerlines of $G(Z)$ and $T(X)$. We apply Hungarian matching to find the corresponding centerlines between the autoencoder output and the initial estimate. Therefore, we treat the initial estimates as labels. We apply thresholding on the existence probability output of $T(X)$ and only use the ``existing'' centerlines in the Hungarian matching.

The initial value of the latent code is obtained from feeding the autoencoder's encoder the outputs of the observation network $T(X)$. We then apply standard gradient descent on the latent code ($Z$) with the loss function given in expression \eqref{eq:second-eq}. While it is possible to use more sophisticated second order optimization techniques, we opted for gradient descent for its ease of implementation and speed. Fig.~\ref{fig:opt} visualizes the process where an initial estimate is mapped to the latent space and the optimization modifies the latent code towards the higher probability region of the prior while faithfully representing the initial estimate.

\subsection{Uncertainty Measurement}

The proposed formulation aims to discover a latent code of the autoencoder such that the output of an estimation network is faithfully reconstructed by the autoencoder as shown in Eq \ref{eq:second-eq}. As mentioned before, the reconstruction error of the autoencoders is a commonly used uncertainty estimator. Therefore, in Eq \ref{eq:second-eq}, we try to find a latent vector that minimizes the uncertainty of the autoencoder subject to the regularization on the norm of the latent code with the initial conditions supplied by an external lane graph estimation network. Then, we can simply use the final reconstruction error of the latent code optimization as our uncertainty estimate of the overall lane graph estimate. The final latent code obtained after the optimization procedure is the most ''sure'' approximation of the autoencoder to the external lane graph estimate. To be precise, the reconstruction loss we use is the L1 loss between the control points of the input lane graph and the reconstructed lane graph. Then the reconstruction error for a sample is given by

\begin{equation}
    |G(Z) - T(X)|; \quad Z = \argmin_Z \quad L_M(G(Z), T(X)) + \alpha||Z||_2  \,,
    \label{eq:uncert}
\end{equation}

Eq \ref{eq:uncert} indicates that the final latent vector obtained through optimization procedure is used to construct a lane graph via the autoencoder decoder. The L1 loss between the control points of this constructed lane graph and the original estimate $T(X)$ is used as the uncertainty measure.

\section{Experiments}

We use the NuScenes \cite{nuscenes2019} and Argoverse \cite{DBLP:conf/cvpr/ChangLSSBHW0LRH19} datasets. The data processing and the train/test splits are the same as~\cite{Can_2021_ICCV}. 

\noindent\textbf{Implementation.}
The BEV area that our method outputs the lane graph is from -25 to 25m in x-direction and 1 to 50m in z direction with a 25cm resolution. We opt for using three Bezier control points. Our implementation is in Pytorch and runs with 7FPS. We set the latent space optimization loss trade-off parameter to $\alpha=0.02$. We run 600 iterations for the latent space optimization with a learning rate of 0.1. We use 32 helper vectors and 100 query vectors. The latent space of the WAE, and the hidden dimension of the transformer is 512. For the training of the WAE, $\beta=2$, $\gamma=1$ and $\theta=1$.

\noindent\textbf{Baselines.}
We compare against state-of-the-art~\textbf{transformer} and~\textbf{Polygon-RNN} based methods proposed in~\cite{Can_2021_ICCV} as well as another baseline which uses the method~\textbf{PINET}~\cite{DBLP:journals/corr/abs-2002-06604} to extract lane boundaries. We only train our method using the transformer based estimates. However, our method can be used in conjunction with any method.

\begin{figure*}
    \centering
    \includegraphics[width=.9\linewidth]{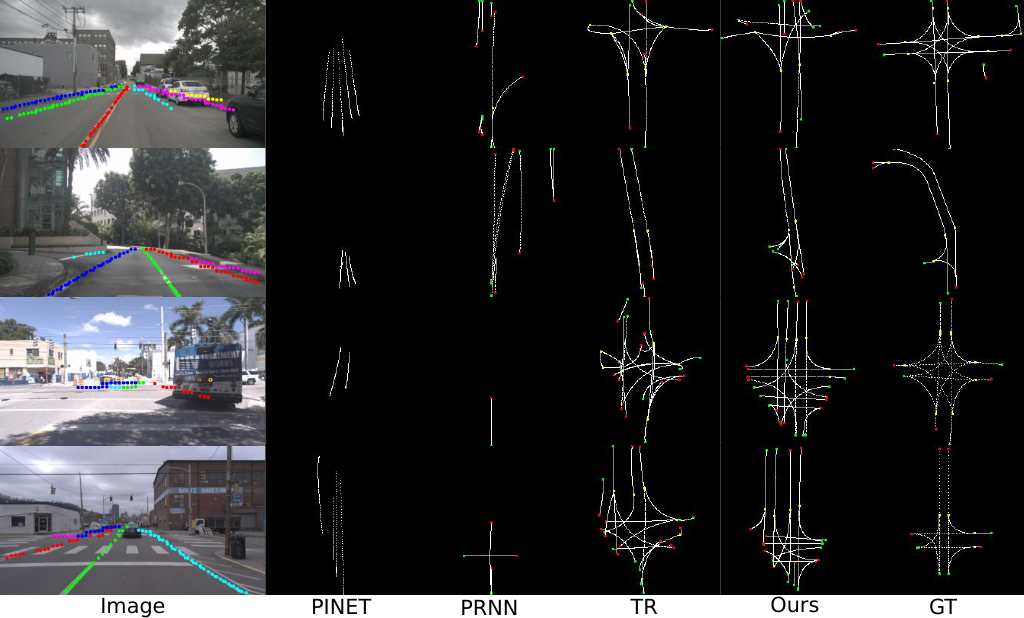}
    \vspace{-1em}
    \caption{The visual results from Nuscenes (top two rows) and Argoverse (bottom two rows). The visuals confirm that the proposed method produces more realistic and accurate lane graphs. The image level lane boundary estimates of PINET are projected onto the image.
    }
    \label{fig:results}
    \vspace{-1em}
\end{figure*}

\noindent\textbf{Ablations.} We trained an autoencoder with the true lane graphs, refereed as \textbf{OursTrue}, instead of the proposed method where the autoencoder is trained with the estimated lane graphs (\textbf{Ours}), see Table~\ref{tab:nusc}. For both networks, we report the performance when no latent space optimization is carried out (\textbf{w/o Opt}) and when we apply latent space optimization. Moreover, we report the convergence behaviour of the latent space optimization for both the true lane graph trained network and ours (Fig.~\ref{fig:graphs}).

\section{Results}

\begin{figure*}
    \centering
    \includegraphics[width=\linewidth]{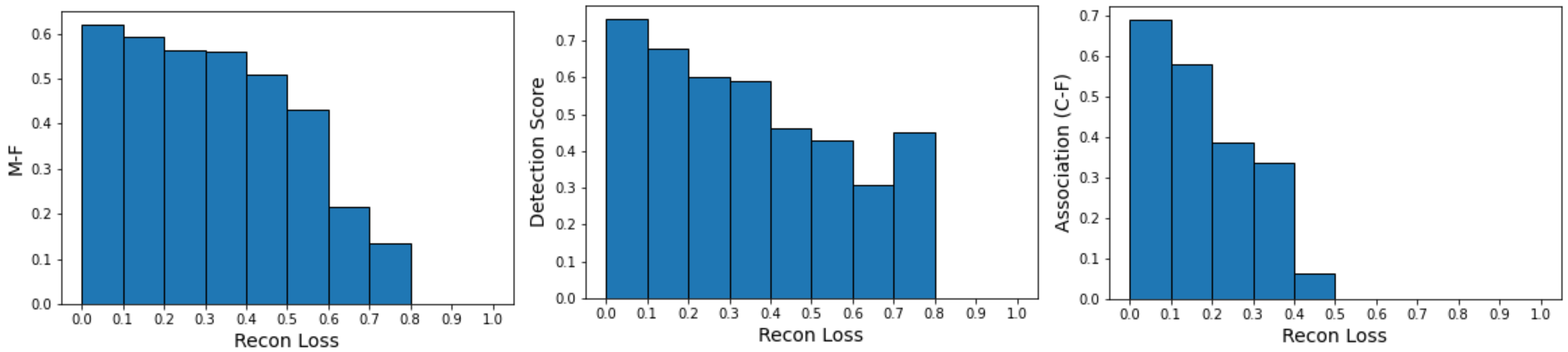}
    \caption{The test-time reconstruction loss versus the metrics are shown. The results indicate that as the reconstruction loss increases, the metrics decrease. Thus, the test time reconstruction loss can be used as an estimate of the performance of the method on a given sample.
    }
    \label{fig:uncer}

\end{figure*}

Here we report the results of our extensive experiments. We first report our methods' performance against the state-of-the-art and then present ablation studies.

\subsection{Comparison with State-of-the-Art}

In Table~\ref{tab:nusc}, quantitative results are given. The first observation from the table is that our method provides a significant performance boost compared to the baseline method of~\cite{Can_2021_ICCV}. Even without latent optimization, \textbf{Ours} provides better results than the baseline. This validates the effectiveness of the proposed novel architecture. Moreover, comparing \textbf{OursTrue} with \textbf{Ours}, it can be seen that the proposed method of using estimated results to train an autoencoder is superior than using the ground truth lane graphs. The final observation is that the proposed latent space optimization steadily improves the results in all metrics, proving the validity of the proposed method. In Table \ref{tab:argo}, the results for Argoverse dataset is given. The results confirm the observations made in Nuscenes dataset results. The proposed optimization procedure consistently improves upon the baselines. Moreover, in training of the autoencoder, using estimated lane graphs produce better results.











\begin{table}[h]
\begin{center}

\begin{tabular}{ |c|c|c|c| }
\hline
& \multicolumn{3}{|c|}{NuScenes}  \\
\hline
\textbf{Method} & M-F &  Detect & C-F \\
\hline
\hline
PINET~\cite{DBLP:journals/corr/abs-2002-06604}  & 49.5 & 19.2 & -  \\

PRNN & 52.9 & 40.5 & 24.5 \\

TR \cite{Can_2021_ICCV}& 56.7  & 59.9 & 55.2\\

\textbf{OursTrue w/o Opt} & 55.8 & 61.8 & 56.3 \\

\textbf{OursTrue} & 56.8 & \textbf{61.9} & 56.5 \\

\textbf{Ours w/o Opt} & 56.0 & 61.1 & 62.9 \\

\textbf{Ours} &  \textbf{57.0} & 61.2 & \textbf{62.9} \\

\hline
\end{tabular}

\end{center}
\vspace{-1em}
\caption{Results on NuScenes dataset. The proposed method significantly outperforms the original method of \cite{Can_2021_ICCV}. Moreover, optimization leads to clearly better results.}
\vspace{-2em}
\label{tab:nusc}
\end{table}

\begin{table}[h]
\begin{center}

\begin{tabular}{ |c|c|c|c| }
\hline
&  \multicolumn{3}{|c|}{Argoverse} \\
\hline
\textbf{Method} &  M-F &  Detect & C-F \\
\hline
\hline
PINET~\cite{DBLP:journals/corr/abs-2002-06604}  & 47.2 & 15.1 & -\\

PRNN  & 45.1 & 40.2 & 31.3\\

TR \cite{Can_2021_ICCV} & 55.6 & 60.1 & 54.9\\

\textbf{OursTrue w/o Opt}  & 56.2 & 62.9 & 53.0\\

\textbf{OursTrue} &  57.2 & 63.7 & 55.0\\

\textbf{Ours w/o Opt}  & 55.2 & 62.4 & 56.9\\

\textbf{Ours} &    \textbf{57.2} & \textbf{63.7} & \textbf{57.2}  \\

\hline
\end{tabular}

\end{center}
\vspace{-1em}
\caption{Results on Argoverse datasets. The proposed method improves upon the existing methods in the literature. It can also be seen that training with the estimated lane graph performs better than using the GT lane graphs for training.}
\vspace{-2em}
\label{tab:argo}
\end{table}

In Fig.~\ref{fig:graphs}, the performance of the \textbf{OursTrue} and \textbf{Ours} methods in all metrics and datasets are given as a function of the iteration. The overall observation is that the latent space optimization produces much better results. It can also be seen that the method converges very quickly and obtaining even higher FPS than reported is possible with a slight decrease in performance. Another observation is the stability in connectivity results throughout the iterations. This indicates the proposed architecture can produce the correct connections directly. Note that the proposed loss for the latent space optimization does not have a connectivity term. The direct optimization over control points of the Bezier curves shows its effect and the mean-f score is consistently improving throughout the optimization process. Improvement in Bezier control point accuracy boosts the other metrics as well since they depend on the centerline matchings between estimated and GT lane graphs.

The visual results are given in Fig.~\ref{fig:results}. While the superior performance of the proposed method is evident, the main improvement comes from improved connectivity. The baseline methods estimate inaccurate connectivity and this results in unrealistic lane graphs. Our method excels in connectivity because the latent vector that produces the lane graph is minimizes a regularized loss. In Argoverse results, our method produces more lanes than the true graph. This results from the input estimates that are supplied from the baseline method. Our proposed method rearranges this inaccurate estimate to produce a better lane graph. In Nuscenes, our method produces more realistic and accurate graphs. In row 2 of the figure, the baseline method estimates multiple centerlines on top of each other and the proposed method untangles them into a logical and accurate lane graph.

\begin{figure}
    \centering
    \includegraphics[width=.8\linewidth]{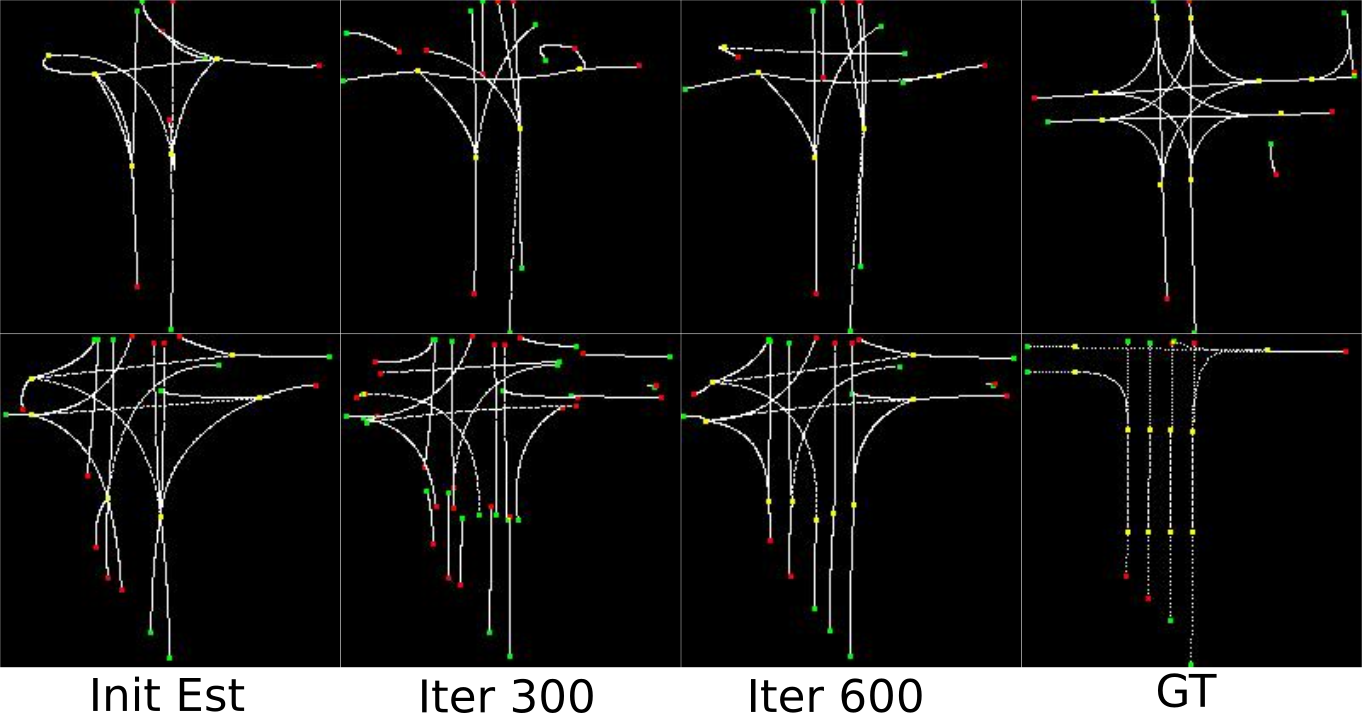}
    \caption{The results of the refinement procedure in different steps: the initial estimate that is fed to the autoencoder, the results after 300 iterations and 600 iterations. Latent space optimization improves the lane graph throughout the refinement process. 
    }
    \label{fig:iterative}

\end{figure}

The visuals for the lane graph refinement results in different iterations of the latent space optimization is given in Fig \ref{fig:iterative}. Compared to initial estimates provided to the refinement process, it can be seen that the lane graph results are more realistic and accurate as more iterations are applied.

\subsection{Uncertainty Estimates}

In order to show the viability of the proposed uncertainty estimation framework, we report the reconstruction error versus the metrics on the test set. We create a histogram for each metric where the mean performance among the scenes that fall in each bin of the histogram is reported as shown in Fig \ref{fig:uncer}. As the reconstruction error of a scene increases, the performance in all metrics decreases. 

From the histograms, it can also be observed that the association measure C-F falls of significantly faster than mean F score and detection ratio. This validates the quantitative results reported in Tables \ref{tab:nusc} and \ref{tab:argo} where the proposed methods perform better than the SOTA especially in association metric. The optimization of latent code of the autoencoder minimizes the reconstruction loss and produces better performance in association. Moreover, the reconstruction loss versus metrics graphs demonstrate the uncertainty for a given sample has to be investigated different for the vertices and the edges of the lane graph.



\section{Conclusion}

In this work we address the task of extracting a complete lane graph from a single onboard camera image. We propose a novel transformer based Wasserstein Autoencoder that can process a lane graph and produce a latent code. The latent code is then used by the decoder to produce the estimated lane graph. In order to utilize this autoencoder, we propose a novel formulation and an accompanying latent space optimization which takes an initial lane graph estimate as input and refines it. The latent space optimization is carried out on the latent code at test time and it does not calculate gradients over the network weights, nor updates them. The results show that both the proposed architecture and the latent space optimization improves the performance significantly. Our extensive ablation studies confirm that optimization is required to achieve the full performance boost, while even without optimization, the proposed architecture can achieve SOTA results. The proposed architecture and formulation have the potential to be used in vision problems that require processing or generating graphs such as scene parsing and layout estimation.





\section{Limitations}

The online test time optimization of the latent vector adds to computational time. However, the batch processing of the samples substantially increases the efficiency of this process.

{\small
\bibliographystyle{ieee_fullname}
\bibliography{egbib}
}

\end{document}